\documentclass[10pt,twocolumn,letterpaper]{article}

\usepackage{subcaption}
\usepackage{cvpr}
\usepackage{times}
\usepackage{epsfig}
\usepackage{graphicx}
\usepackage{amsmath}
\usepackage{amssymb}
\usepackage{pifont}
\usepackage{dsfont}

\newcommand{\myparagraph}[1]{\vspace{8pt}\noindent\textbf{#1}}

\DeclareMathOperator{\GBN}{GBN}

\DeclareMathOperator{\DA}{DA_{BN}}

\newcommand{\set}[1]{\mathcal{#1}}
\usepackage[pagebackref=true,breaklinks=true,letterpaper=true,colorlinks,bookmarks=false]{hyperref}

 \cvprfinalcopy % *** Uncomment this line for the final submission

\def\cvprPaperID{5575} % *** Enter the CVPR Paper ID here

\ifcvprfinal\fi
\begin{document}

%%%%%%%%% TITLE
\title{AdaGraph: Unifying %Merging
Predictive and Continuous \\ Domain Adaptation through Graphs}

\author{Massimiliano Mancini$^{1,2}$, Samuel Rota Bul\`o$^3$, Barbara Caputo$^{4,5}$, Elisa Ricci$^{2,6}$\\
$^1$Sapienza University of Rome, $^2$Fondazione Bruno Kessler, $^3$Mapillary Research,\\ $^4$Politecnico di Torino,$^5$Italian Institute of Technology, $^6$University of Trento\\
{\tt\small mancini@diag.uniroma1.it},{\tt\small samuel@mapillary.com},{\tt\small barbara.caputo@polito.it},{\tt\small eliricci@fbk.eu}
}

\maketitle
%\thispagestyle{empty}

%%%%%%%%% ABSTRACT
\begin{abstract}
The ability to categorize is a cornerstone of visual intelligence, and a key functionality for artificial, autonomous visual machines. This problem will never be solved without algorithms able to adapt and generalize across visual domains.
%, i.e. to recognize reliably object classes imaged in diverse conditions, with different styles, without the need of costly re-training, for which often there is no annotated data available. 
Within the context of domain adaptation and generalization, this paper focuses on the \emph{predictive domain adaptation} scenario, namely the case where no target data are available and the system has to learn to generalize from annotated source images plus unlabeled samples with associated metadata from auxiliary domains.
Our contribution is the first deep architecture that tackles predictive domain adaptation, able to leverage over the information brought by the auxiliary domains  through a graph. Moreover, we present a simple yet effective strategy that allows us to take advantage of the incoming target data at test time, in a continuous domain adaptation scenario. Experiments on three benchmark databases support the value of our approach.

\end{abstract}

%%%%%%%%% BODY TEXT
\section{Introduction}

Over the past years, deep learning has enabled rapid progress in many visual recognition tasks, even surpassing human performance \cite{ILSVRC15}. While deep networks exhibit excellent generalization capabilities, previous studies \cite{donahue2014decaf} demonstrated that their performance drops when test data significantly differ from training samples.
In other words
%, as any other supervised learning approach, 
deep models %, by assuming that training and test data are drawn from the same underlying distribution, 
suffer from the \textit{domain shift} problem, \textit{i.e.} classifiers trained on \textit{source} data do not perform well when tested on samples in the \textit{target} domain. In practice, domain shift arises in many computer vision tasks, as many factors (\eg lighting changes, different view-points, \etc) determine appearance variations in visual data. 

To cope with this, several efforts focused on developing Domain Adaptation (DA) techniques \cite{wang2018deep},
 attempting to
reduce the mismatch between source and target data distributions to learn accurate prediction models for the target domain. In the challenging case of unsupervised DA, only source data are labelled while no annotation is provided for target samples. 
%Previous work can be 
%Over the last few years several deep learning-based approaches have been proposed for unsupervised DA which can be 
%roughly divided in three main categories. 
%The first includes approaches modeling the source and target data distributions in term of their first or second order statistics, trying to match them \cite{long2016deep,morerio2017minimal,carlucci2017autodial,french2018self,mancini2018boosting}. The second group of methods define specific loss functions which promote the learning of domain invariant deep representations \cite{tzeng2015simultaneous,ganin2014unsupervised}. Other approaches are derived from Generative Adversarial Networks (GANs)\cite{Goodfellow:GAN:NIPS2014} and exploit the idea of translating images from one domain to another for the purpose of DA \cite{bousmalis2016domain,russo17sbadagan,hoffman2017cycada}.
Although it might be reasonable for some applications to have target samples available during training,
%
%A common assumption for all unsupervised domain adaptation methods is that target data are available during training.
%While this is reasonable in some applications, 
it is hard to imagine that we can collect data for every possible target. More realistically, we aim for prediction models which can generalize to \textit{new, previously unseen target domains}. Following this idea, previous studies proposed the Predictive Domain Adaptation (PDA) scenario \cite{yang2016multivariate}, where  neither the data, nor the labels from the target are available during training. Only annotated source samples are available, together with additional information from a set of auxiliary domains, in form of unlabeled samples and associated metadata (\textit{e.g.} corresponding to the image timestamp or to camera pose, etc).

\begin{figure}[t]
\begin{center}
   \includegraphics[width=1.0\columnwidth]{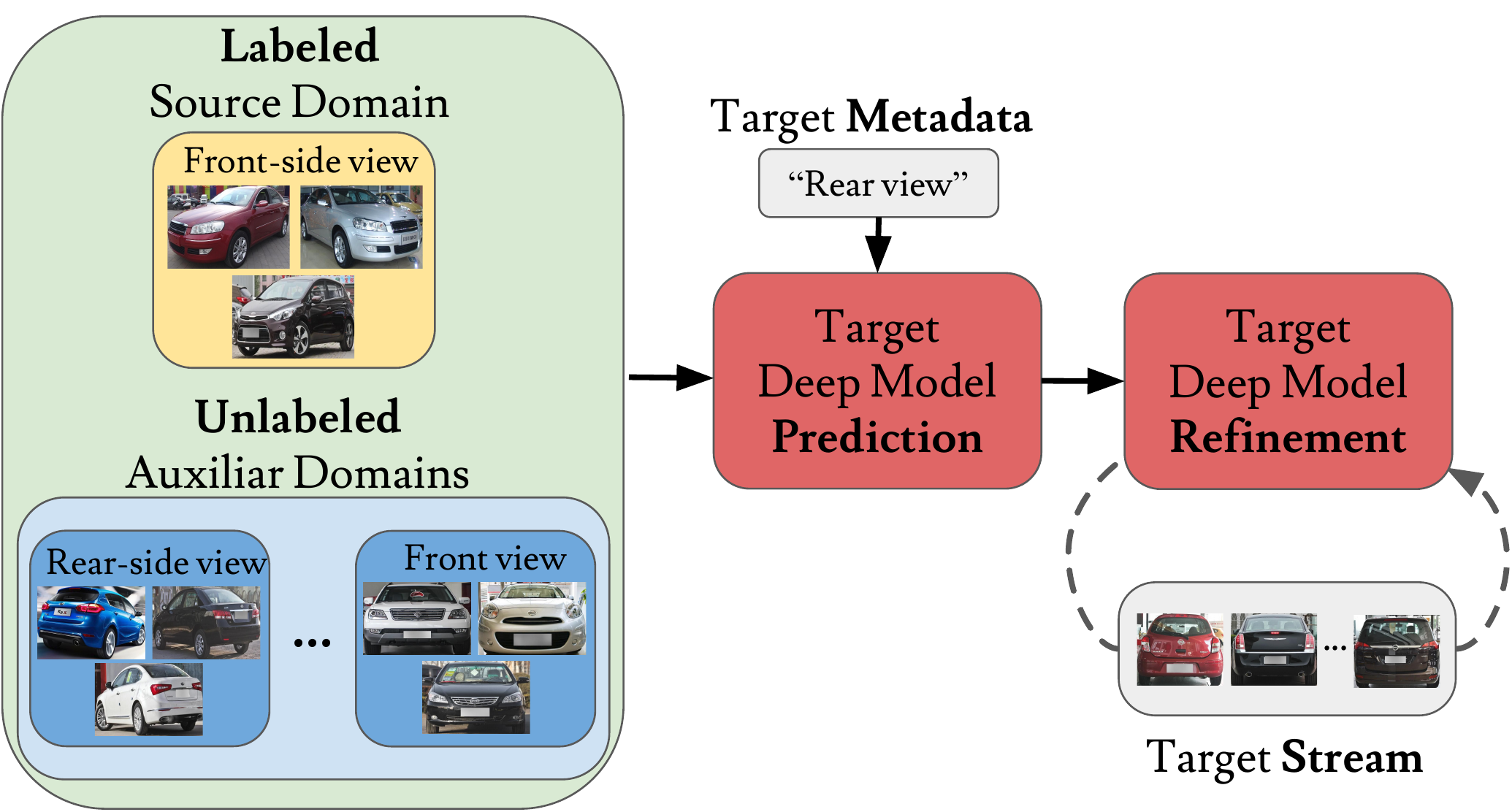}
\end{center}
\vspace{-5pt}
   \caption{Predictive Domain Adaptation. During training we have access to a labeled source domain (yellow block) and a set of unlabeled auxiliary domains (blue blocks), all with associated metadata. At test time, given the metadata corresponding to the unknown target domain, we predict the parameters associated to the target model. %deep model for that domain. 
   This predicted model is further refined during test, while continuously receiving data of the target domain. Best viewed in color.}
\vspace{-15pt}
\label{fig:teaser}
\end{figure}

In this paper we introduce a deep architecture for PDA.
Following recent advances in DA \cite{carlucci2017autodial,li2018adaptive,mancini2018boosting}, we propose to learn a set of domain-specific models by considering a common backbone network with domain-specific alignment layers embedded into it. We also propose to exploit metadata and auxiliary samples by
 building a graph which explicitly describes the dependencies among domains. 
Within the graph, nodes represent domains, while edges encode relations between domains, imposed by their metadata. %We map each domain to a subset of the network parameters and there exist several auxiliary domains described in the same parametric space.
Thanks to this construction,  when metadata for the target domain are available at test time, the domain-specific model can be recovered. We further exploit target data directly at test time by devising an approach for continuously updating the deep network parameters once target samples are made available {(Figure \ref{fig:teaser})}.  We demonstrate the effectiveness of our method with experiments on three datasets: the \textit{Comprehensive Cars} (CompCars) \cite{yang2015large}, the \textit{Century of Portraits} \cite{ginosar2015century}  and the \textit{CarEvolution} datasets \cite{RematasICCVWS13}, showing that our method outperforms state of the art PDA approaches. Finally, we show that the proposed approach for continuous updating of the network parameters can be used for continuous domain adaptation, producing more accurate predictions than previous methods \cite{hoffman2014continuous,li2018domain}.

\textbf{Contributions.} To summarize, the contributions of this work are: (i) we propose the first deep architecture for addressing the problem of PDA; (ii) we present a strategy for injecting metadata information within a deep network architecture by encoding the relation between different domains through a graph; (iii) we propose a simple strategy for refining the predicted target model which exploits the incoming stream of target data directly at test time.
%\begin{itemize}
 %   \item The first deep architecture addressing the PDA problem.
  %  \item A way to couple a deep architecture with a graph which allows to (i) map network parameters to external information, in or case domain metadata (ii) combine and propagate network parameters based on the position of a node within the graph.
   % \item A simple strategy for continuous update of model parameter which is beneficial not only for PDA but which, used in isolation, leads to sota in continuous domain adaptation
%\end{itemize}
%(i) we present the first deep architecture addressing PDA, (ii) , (iii) a simple strategy for continuous update of model parameter which is beneficial not only for PDA but which, used in isolation, leads to sota in contiuous domain adaptation.

%\massi{DISCLAIMER Dataset e Timothy}

%-------------------------------------------------------------------------

\section{Related Works}
%In this section we review previous works on unsupervised domain adaptation, with special emphasis on methods considering deep networks and multiple domains and approaches that not require target data.
\vspace{-5pt}
\textbf{Unsupervised Deep Domain Adaptation.} Previous works on deep DA learn domain invariant representations by exploiting different architectures, such as Convolutional Neural Networks  \cite{long2015learning,tzeng2015simultaneous,ganin2014unsupervised,carlucci2017autodial,li2018adaptive,french2018self,LOAD_ICRA}, deep autoencoders \cite{zeng2014deep} or GANs \cite{russo17sbadagan,hoffman2017cycada}. Some methods describe source and target features distributions considering their first and second order statistics and minimize their distance either defining an appropriate loss function \cite{long2015learning} or deriving some domain alignment layers \cite{li2018adaptive,carlucci2017autodial, french2018self}. Other approaches
rely on adversarial loss functions \cite{tzeng2015simultaneous,ganin2016domain} to learn domain agnostic representations. GAN-based techniques \cite{bousmalis2016domain,hoffman2017cycada,russo17sbadagan} for unsupervised DA focus directly on images and aim at generating either target-like source images or source-like target images. Recent works also showed that considering both the transformation directions is highly beneficial \cite{russo17sbadagan}. 

In many applications multiple source domains may be available. This fact has motivated the study of multi-source DA algorithms \cite{xu2018deep,mancini2018boosting}. In  \cite{xu2018deep} an adversarial learning framework for multi-source DA is proposed, inspired by \cite{ganin2014unsupervised}. A similar  adversarial strategy is also exploited in \cite{MDAN_ICLRW18}. In \cite{mancini2018boosting}
a deep architecture is proposed to discover multiple latent source domains in order to improve the classification accuracy on target data.

Our work performs domain adaptation by embedding into a deep network domain-specific normalization layers as in \cite{li2018adaptive,carlucci2017autodial,french2018self,roy2019unsupervised}. However, the design of our layers is different as they are required to guarantee a continuous update of parameters and to exploit information from the domain graph. Our approach considers information from multiple domains at training time. However, instead of having labeled data from all source domains, we do not have annotations for samples of auxiliary data.

{Finally, our work is linked to graph-based domain adaptation methods \cite{ding2018graph,das2018graph}. Differently from these works however, in our approach a node does not represent a single sample but a whole domain and edges do not link semantically related samples but domains with related metadata.}

\vspace{-5pt}
\myparagraph{Domain Adaptation without Target Data.} 
%\massi{Aggiungere 2 righe su lavori di Van Gool chiesti da R1?}
In some applications, the assumption that target data are available during training does not hold. This calls for DA methods able to cope with the domain shift by exploiting either the stream of incoming target samples, or side information describing possible future target domains. 

The first scenario is typically referred to as {continuous} \cite{hoffman2014continuous} or {online} DA \cite{mancini2018kitting}. To address this problem, in \cite{hoffman2014continuous} a manifold-based DA technique is employed to model an evolving target data distribution. In \cite{li2018domain} Li \etal propose to sequentially update a low-rank exemplar SVM classifier as data of the target domain becomes available. In \cite{lampert2015predicting}, the authors propose to extrapolate the target data dynamics within a reproducing kernel Hilbert space. 

The second scenario corresponds to the problem of predictive DA tackled in this paper. PDA is introduced in \cite{yang2016multivariate}, where a multivariate regression approach is described for learning a mapping between domain metadata and points in a Grassmanian manifold. Given this mapping and the metadata for the target domain, two different strategies are proposed to infer the target classifier. %the subspace in the manifold relative to the metadata itself involving \textit{Direct} kernel regression  or an \textit{Indirect} method. The Geodesic Flow Kernel \cite{gong2012geodesic} algorithm is used to produce a model for the target domain exploiting the given source.}}

Other closely related tasks are the problems of zero shot domain adaptation and domain generalization. 
In zero-shot domain adaptation (ZDDA) \cite{peng2018zero} the task is to learn a prediction model in the target domain 
under the assumption that task-relevant source-domain data and task-irrelevant
dual-domain paired data are available. {We highlight that the PDA problem is related, but different, from ZDDA.
%Both problems address DA when no target data is available during training. However, 
ZDDA assumes that the domain shift is known during training from the presence of data of a different task but with the same visual appearance of source and target domains, % of task-irrelevant source-target pairs, 
while in PDA metadata of auxiliary domains is the only available information, and the target metadata is received \textit{only at test time}. % are used with no information about the target domain during training. 
For this reason, ZDDA is not applicable to a PDA scenario, and it cannot predict the classification model for a target domain given only the metadata.}

Domain generalization methods \cite{muandet2013domain,li2017deeper,Antonio_GCPR18,doretto2017} attempt to learn domain-agnostic classification models by exploiting labeled source samples from multiple domains but without having access to target data. Similarly to Predictive DA in domain generalization, multiple datasets are available during training. However, in PDA data from auxiliary source domains are not labeled. %nly 1 of them is actually labeled. 

\section{Method}
\subsection{Problem Formulation} 
\vspace{-5pt}
\label{sec:problem-formulation}
Our goal is to produce a model that is able to accomplish a task in a \textit{target} domain $\mathcal{T}$ for which \textit{no data} are available during training, neither labeled nor unlabeled. The only information we can exploit is a characterization of the content of the target domain in the form of \textit{metadata} $m_{\set T}$ plus a set of known domains $\mathcal{K}$, each of them having associated metadata. 
All domains in $\mathcal{K}$ carry information about the task we want to accomplish in the target domain. In particular, since in this work we focus on classification tasks, we assume that images from the domains in $\mathcal{K} $ and $\mathcal{T}$ can be classified with semantic labels from a same set $\mathcal{Y}$. 
As opposed to standard DA scenarios, the target domain $\mathcal T$ does not necessarily belong to the set of known domains $\mathcal{K}$. Also, we assume that $\mathcal{K}$ can be partitioned into a \textit{labeled} \textit{source} domain $\mathcal{S}$ and $N$ \textit{unlabeled} \textit{auxiliary} domains $\mathcal{A}=\{A_1,\cdots,A_N\}$. 

In the specific, this paper focuses on predictive DA (PDA) problems aimed at regressing the target model parameters using data from the domains in $\mathcal K$. We achieve this objective by (i) interconnecting each domain in $\mathcal{K}$ using the given domain metadata; (ii) building domain-specific models from the data available in each domain in $\mathcal{K}$;  (iii) exploiting the connection between the target domain and the domains in $\mathcal{K}$, inferred from the respective metadata, to regress the model for $\mathcal{T}$.
%. In particular, in this work we focus on classification tasks, thus our goal is to associate to an input image $x$ its correct semantic label $y\in\mathcal{Y}$, where we assume $\mathcal{Y}$ to be a closed set. To produce this classification model, 
%To produce this classification model,we have access to a labeled \textit{source} domain $\mathcal{S}=\{(x^s_1,y^\mathcal{S}_1),\cdots,(x^s_N,y^s_N)\}$, where $x^\mathcal{S}_i \in \mathcal{X}$ and $y^\mathcal{S}_i \in \mathcal{Y}$ its associated semantic label. However, opposite to standard DA scenarios, we have \textit{no data} of the target domain available during training, neither labeled nor unlabeled. In order to cope with the domain shift between the two domains, 
%The goal of predictive domain adaptation is to produce a model able to address a given task for a target domain using just the target domain meta information plus a set of auxiliary domains associated with metadata.

A schematic representation of the method is shown in Figure 2. We propose to use a graph because of its seamless ability to encode relationships within a set of elements (domains in our case). Moreover, it  can be easily manipulated to include novel elements (such as the target domain $\mathcal{T}$).

\subsection{AdaGraph: Graph-based Predictive DA}
\vspace{-5pt}
\label{sec:graph-da}

We model the dependencies between the various domains by instantiating a graph composed of nodes and edges. Each node represents a different domain and each edge measures the relatedness of two domains.  Each edge of the graph is weighted, and the strength of the connection is computed as a function of the domain-specific metadata. At the same time, in order to extract one model for each available domain, we employ recent advances in domain adaptation involving the use of domain-specific batch-normalization layers \cite{li2018adaptive,carlucci2017just}.  With the domain-specific models and the graph we are able to predict the parameters for a novel domain that lacks data by simply (i) instantiating a new node in the graph and (ii) propagating the parameters from nearby nodes, exploiting the graph connections.

\vspace{-5pt}
\myparagraph{Connecting domains through a graph.}
Let us denote the space of domains as $\mathcal{D}$ and the space of metadata as $\mathcal{M}$. As stated in Section \ref{sec:problem-formulation}, in the PDA scenario, we have a set of known domains $\mathcal{K}=\{k_1,\cdots,k_n\}\subset\mathcal{D}$ and a bijective mapping $\phi:\mathcal{D}\mapsto\mathcal{M}$ relating domains and metadata. % associating to a domain $d\in\mathcal{D}$ a metadata $m\in\mathcal{M}$. 
%This mapping is injective in the subspace $\mathcal{K}$ since there may exist a metadata $m\in\mathcal{M}$ to which no domain is associated. 
For simplicity, we regard as \textit{unknown} some metadata $m$ that is not associated to domains in $\set K$, \ie such that $\phi^{-1}(m)\notin \set K$.
%\begin{align}
 %   \forall& k\in \mathcal{K}, \; \exists \, m\in\mathcal{M}:\,m=\phi(k) \\
  %  \forall& k_i,k_j\in \mathcal{K}\;\;  \phi(k_i)=\phi(k_j) \iff k_i=k_j 
%\end{align}
%We want to highlight that there may exist a metadata $m\in\mathcal{M}$ to which no domain is associated (\ie $\nexists k\in \mathcal{K} : m=\phi(k)$). 

In this work we structure the domains as a \textit{graph} $\mathcal{G}=(\mathcal{V},\mathcal{E})$, where $\set V\subset\set D$ represents the set of vertices corresponding to domains and $\set E\subseteq\set V\times\set V$ the set of edges, \ie relations between domains. Initially the graph contains only the known domains so $\set V=\set K$. In addition, we define an edge weight $\omega:\set E\to\mathbb R$ that measures the relation strength between two domains $(v_1,v_2)\in \set E$ by computing a distance between the respective metadata, \ie
\begin{equation}
    \label{eq:edges}
    \omega(v_1,v_2)=e^{-d(\phi(v_1),\phi(v_2))}\,,
\end{equation}
where $d:\set M^2\to\mathbb R$ is a distance function on $\set M$.
%Exploiting the set $\mathcal{K}$ of known domains, we define a graph $\mathcal{G}=(\mathcal{V},\mathcal{E})$ where a node $v\in\mathcal{V}$ is linked to a known domain %$\mathcal{V}\equiv\mathcal{K}$ 
%by $\omega: \mathcal{K} \mapsto \mathcal{V}$ and a metadata by $\phi^\prime: \mathcal{V} \mapsto \mathcal{M}$.  $\mathcal{E}$ is the set of edges connecting the domains, with $e_i^j\in \mathcal{E}$ denoting the weighted edge connecting domain $k_i$ to domain $k_j$.  %Since the set of nodes and known domains coincide, in the following we will refer to the set of nodes and known domains interchangeably.
%A domain $k$ and its linked node share the same metadata \ie $\phi(k)=\phi^\prime(\omega(k))$.

%To estimate the weights of each edge we exploit the metadata linked to the nodes. In particular we define the edge between two domains $k,j \in \mathcal{K}$ as follows:
%\begin{equation}
%    \label{eq:edges}
%    e_v^z=\exp \big(-d(v,z)\big), \;\;\;\text{with}\;\;\;\omega(k)=v \text{ and } \omega(j)=z
%\end{equation}
%where $d(\cdot,\cdot)$ is a distance measure in $\mathcal{M}$. With the defined graph we have characterized the relation between each domain to the other, exploiting the information given by the metadata. 

Let $\Theta$ be the space of possible model parameters and assume we have properly exploited the domain data from each domain in $k\in\mathcal{K}$ to learn a set of domain-specific models (we will detail this procedure in the next subsection). We can then define a mapping $\psi: \mathcal{K}\mapsto\Theta$, relating each domain to its set of domain-specific parameters. %This mapping consequently relates a node $v\in \mathcal{V}$ to its set of parameters. In particular, defining as $\psi^\prime: \mathcal{V}\mapsto\Theta$ we have:
%\begin{equation}
%\label{eq:node-domain-param-mapping}
%    \forall v\in\mathcal{V}, k\in\mathcal{D} \,\,\, \psi^\prime(v)=\psi(k)\iff\omega(k)=v 
%\end{equation} 
%which means that a known domain and its associated node share the same set of parameters. 
Given some metadata $m\in\set M$ we can recover an associated set of parameters via the mapping $\psi\circ\phi^{-1}(m)$ provided that $\phi^{-1}(m)\in\set K$.
%Notice that, with $\psi$, we are able to directly obtain the domain parameters from a metadata $m\in\mathcal{M}$ only if $m$ is in the codomain of $\phi$ \ie the metadata is associated with one of the known domains. 
%What we would like to have instead, is the possibility of producing an inverse mapping of $\phi$ even for unknown metadata, such as the target domain one $m_{\set T}$. %, which do not correspond to any domain in $\mathcal{K}$.
%The problem of $\psi$ is that it is only define for domains $k\in\mathcal{K}$ and it is not able to directly provide us the set of model parameters relative to a metadata $m_{\set T}$ for which $\nexists k\in\mathcal{K} : \phi(k)=m_{\set T}$.
In order to deal with metadata that is unknown, we introduce the concept of \textit{virtual} node. Basically, a virtual node $\tilde{v} \in \mathcal{V}$ is a domain for which no data are available but we have metadata $\tilde m$ associated to it, namely $\tilde{m}=\phi(\tilde v)$. For simplicity, let us directly consider the target domain $\mathcal{T}$. We have $\mathcal{T}\in \mathcal{D}$ and we know $\phi(\mathcal{T})=m_{\set T}$. Since no data of $\mathcal{T}$ are available, we have no parameters that can be directly assigned to the domain. However, we can estimate parameters for $\set T$ by using the domain graph $\mathcal{G}$. Indeed, we can relate $\set T$ to other domains $v\in\set V$ using $\omega(\mathcal{T},v)$ defined in \eqref{eq:edges} by opportunely extending $\set E$ with new edges $(\set T,v)$ for all or some $v\in\set V$ (\eg we could connect all $v$ that satisfy $\omega(\mathcal{T},v)>\tau$ for some $\tau$). The extended graph $\set G'=(\set V\cup\{\set T\},\set E')$ with the additional node $\set T$ and the new edge set $\set E'$ can then be exploited to estimate parameters for $\set T$ 
by \textit{propagating} the model parameters from nearby domains. Formally we regress the parameters $\hat{\theta}_{\mathcal{T}}$ through
\begin{equation}
    \label{eq:regression-metadata}
    \hat{\theta}_{\mathcal{T}}=\psi(\mathcal{T})=\frac{\sum_{(\set T,v)\in \set E'}\omega(\mathcal{T},v)\psi(v)}
    {\sum_{(\set T,v)\in \set E'}\omega(\mathcal{T},v)}\,,
\end{equation}
where we normalize the contribution of each edge by the sum of the weights of the edges connecting node $\mathcal{T}$. %recalling that $\mathcal{T}\in\mathcal{D}$ and $\omega(\mathcal{T})=\tilde{v}$, from \eqref{eq:node-domain-param-mapping} we have that:
%\begin{equation}
 %   \label{eq:target-assignment}
  %  \omega(\mathcal{T})=\tilde{v} \implies \psi(\mathcal{T})=\psi^\prime(\tilde{v})
%\end{equation}
With this formula we are able to provide model parameters for the target domain $\mathcal{T}$ and, in general, for any unknown domain by just exploiting the corresponding metadata.% in order to build the connections with any other node/domain in $\mathcal{G}$. 

We want to highlight that this strategy simply requires extending the graph with a virtual node $\tilde v$ and computing the relative edges. While the relations of $\tilde{v}$ with other domains can be inferred from given metadata, as in \eqref{eq:edges}, there could be cases in which no metadata are available for the target domain. 
In such situations, we can still exploit the incoming target image $x$ to build a probability distribution over nodes in $\mathcal{V}$, in order to assign the new data point to a mixture of known domains. To this end, let use define $p(v|x)$ the conditional probability of an image $x\in\mathcal{X}$, where $\mathcal{X}$ is the image space, to be associated with a domain $v\in\mathcal{V}$. From this probability distribution, we can infer the parameters of a classification model for $x$ through:%Since the metadata $m_{\set T}$ to which $x$ it is likely to be unknown, if we consider only the set of nodes $\mathcal{V}$ belongs could be an unknown metadata,  Moreover, since there may exist no node in $\mathcal{V}$ associated to a metadata $m$, let us define $\mathcal{V}^*=\mathcal{V}\cup\hat{\mathcal{V}}$, where $\mathcal{V}^*$ is the set comprising both the nodes in $\mathcal{V}$ and possible virtual nodes $\hat{\mathcal{V}}$. may correspond to a metadatalet us define $ $We can marginalize over all $\mathcal{M}$ to get an estimate for the parameters of $x$ with:
\begin{equation}
    \label{eq:regression-image}
    \hat{\theta}_x=\sum_{v\in \mathcal{V}}p(v|x)\cdot\psi(v)
\end{equation}
where $\psi(v)$ is well-defined for each node linked to a known domain, while it must be estimated with \eqref{eq:regression-metadata} for each virtual domain $\tilde{v}\in\mathcal{V}$ for which $p(\tilde{v}|x)>0$.
%Moreover, notice that \eqref{eq:regression-image} is a generalization of \eqref{eq:regression-metadata}, since the presence of the metadata constrains $p(v|x)=1$ if $\phi(\tilde{v})=m_{\set T}$ and $0$ otherwise, since $\phi$ is bijective.

In practice, the probability $p(v|x)$ is constructed from a metadata classifier $f^m$, trained on the available data, that provides a probability distribution over $\set M$ given $x$, which can be turned into a probability over $\set D$ through the inverse mapping $\phi^{-1}$.

 \begin{figure*}[t]
\begin{center}
   \includegraphics[width=0.8\linewidth,height=0.335\linewidth]{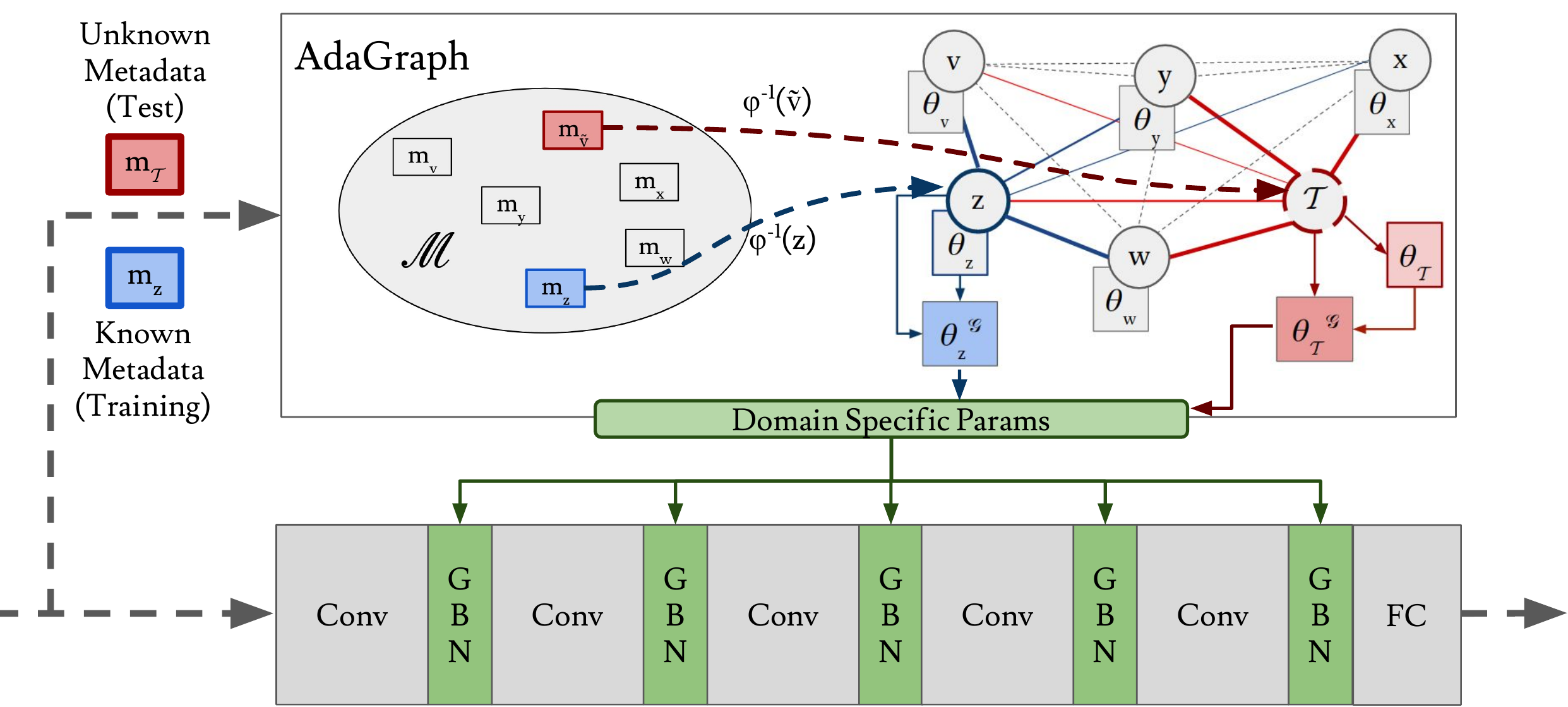}
   \vspace{-8pt}
\end{center}
   \caption{AdaGraph framework (Best viewed in color). Each BN layer is replaced by its GBN counterpart. The parameters used in a GBN layer are computed from a given metadata and the graph. %The relation $\phi^{-1}$ maps a metadata to a node in our graph.
   Each domain in the graph (circles) contains its specific parameters (rectangular blocks). During the training phase (blue part), a metadata (\ie $m_z$, blue block) is mapped to its domain (z). While the statistics of GBN are determined only by the one of $z$ ($\theta_z$), scale and bias are computed considering also the graph edges. During test, we receive the metadata for the target domain ($m_{\set T}$, red block) to which no node is linked. Thus we initialize $\set T$ and we compute its parameters and statistics exploiting the connection with the other nodes in the graph ($\theta_{\set T}$).}
\label{fig:method}
\vspace{-17pt}
\end{figure*}

\vspace{-4pt}
\myparagraph{Extracting node specific models.} 
We have described how to regress model parameters for an unknown domain by exploiting the domain graph.
Now, we focus on the actual problem of training domain-specific models using data available from the known domains $\set K$. 
%Now that we have defined how to regress model parameters from a domain graph, we must explicit how we can actually compute the set of domain/node specific models given the known domains $\mathcal{K}$. 
Since $\mathcal{K}$ entails a labeled source domain $\mathcal{S}$ and a set of auxiliary domains $\mathcal{A}$,  we cannot simply train
independent models with data from each available domain due to the lack of supervision on domains in $\mathcal{A}$ for the target classification task. 
For this reason, we need to estimate the model parameters for the unlabeled domains $\mathcal{A}$ by exploiting DA techniques.

Recent works~\cite{li2018adaptive,carlucci2017autodial,carlucci2017just} have shown the effectiveness of applying domain-specific batch-normalization ($\DA$) layers to address domain adaptation tasks. In particular, these works rewrite each batch-normalization layer~\cite{ioffe2015batch} (BN) of the network in order to take into account domain-specific statistics. %A standard BN layer normalizes its input according to:
%\begin{equation}
 %   \label{eq:BN}
  %  \BN(x,k)=\gamma \cdot \frac{x-\mu}{\sqrt{\sigma^2+\epsilon}} + \beta\,,
%\end{equation}
%where $\mu$ is the estimated mean of $x$, $\sigma^2$ the estimated variance of $x$, $\gamma$ and $\beta$ are learnable scale and bias parameters, respectively, and $\epsilon$ is a small constant used to avoid numerical instabilities. Notice that we have dropped the dependencies on spatial location and channel dimension for simplicity. 
{Given a domain $k$, a $\DA$ layer differs from standard BN by including domain-specific information: 
\begin{equation}
    \label{eq:da-bn}
    \DA(x,k)=\gamma \cdot \frac{x-\mu_k}{\sqrt{\sigma^2_k+\epsilon}} + \beta\,,
\end{equation}
where the mean and variance statistics $\{\mu_k,\sigma_k\}$ are estimated from $x$ conditioned on domain $k$, $\gamma$ and $\beta$ are learnable scale and bias parameters, respectively, and $\epsilon$ is a small constant used to avoid numerical instabilities. Notice that we have dropped the dependencies on spatial location and channel dimension for simplicity.} % \ie $\mu_k=\E_{x\sim k}[x]$ and $\sigma^2_k=\V_{x \sim k}[x]$. 
The effectiveness of this simple DA approach is due to the fact that features of source and target domains are forced to be aligned to the same reference distribution, and this allows to implicitly address the domain shift problem.

In this work we exploit the same ideas to provide each node in the graph with its own BN statistics. At the same time, we depart from \cite{li2018adaptive,carlucci2017just} since we do not keep scale and bias parameters shared across the domains, but we include also them within the set of domain-specific parameters. 
%We limit the domain-specific parameters to the ones pertaining to $\DA$ layers because of the limited number of parameters they require and their easy scalability to multi-domain scenarios. 

In this scenario, the set of parameters for a domain $k$ $\psi(k)=\theta_k$ is composed of different parts. Formally for each domain we have $\psi(k)=\{\theta^{a},\theta^{s}_k\}$, where $\theta^a$ holds the domain-agnostic components and $\theta^s_k$ the domain-specific ones. In our case $\theta^a$ comprises parameters from standard layers (\ie the convolutional and fully connected layers of the architecture), while $\theta^s_k$ comprises parameters and statistics of the domain-specific BN layers. %Notice that, in the following, we will interchangeably refer to a domain $k$ and the node $v=\omega(k)$ since there exists a one to one map in the subspace $\mathcal{K}$ that we use in the training phase.
 
We start by using the labeled source domain $\mathcal{S}$ to estimate $\theta^a$ and initialize $\theta^s_\mathcal{S}$. In particular, we obtain $\theta_{\mathcal{S}}$ %Assuming that $s$ is the node associated to $\mathcal{S}$ (\ie $s=\omega(\mathcal{S})$ we estimate $\theta_s=\psi^\prime(s)$ 
by minimizing the standard cross-entropy loss:
\begin{equation}
    \label{eq:loss-source}
    L(\theta_\mathcal{S}) = -\frac{1}{|\mathcal{S}|} \sum_{(x,y)\in\mathcal{S}} \log (f_{\theta_\mathcal{S}}(y;x)) \,,
\end{equation}
where $f_{\theta_{\mathcal{S}}}$ is the classification model relative to the source domain, with parameters $\theta_\mathcal{S}$. %By minimizing $\eqref{eq:loss-source}$ we are able to estimate not only the set of domain-specific parameters $\theta_s$, but also the set of shared parameters $\theta_a$.

To extract the domain-specific parameters $\theta^s_k$ for each $k \in \mathcal{K}$, we employ 2 steps: the first is a selective forward pass for estimating the domain-specific statistics while the second is the application of a loss to further refine the scale and bias parameters. Formally, we replace each BN layer in the network with a \textit{GraphBN} counterpart (GBN), where the forward pass is defined as follows:
\begin{equation}
  \label{eq:graph-forward-std}
  \GBN(x,v)=\gamma_v \cdot\frac{x-\mu_v}{\sqrt{\sigma_v^2+\epsilon}}\; +\; \beta_v\,.
\end{equation}
Basically in a GBN layer, the set of BN parameters and statistics to apply is conditioned on the node/domain to which $x$ belongs. During training, as for standard BN, we update the statistics by leveraging their estimate obtained from the current batch $\mathcal{B}$:
\begin{equation}
  \label{eq:graph-stats}
  \hat{\mu_v}=\frac{1}{|\mathcal{B}_v|} \sum_{x\in \mathcal{B}_v} x\;\;\;\text{and}\;\;\;
  \hat{\sigma^2_v}=\frac{1}{|\mathcal{B}_v|} \sum_{x\in \mathcal{B}_v} (x-\mu_v)^2\,,
\end{equation}
where $\mathcal{B}_v$ is the set of elements in the batch belonging to domain $v$. As for the scale and bias parameters, we optimize them by means of a loss on the model output. For the auxiliary domains, since the data are unlabeled, we use an entropy loss, while a cross-entropy loss is used for the source domain: %Denoting the classification model $f_{\theta_k}=f_{\psi(k)}$ as $f_k$ for simplicity, we have: 
\begin{equation}
    \label{eq:loss-combined}
    \begin{split}
    L(\Theta^s) =-\frac{1}{|\mathcal{S}|} & \sum_{(x,y)\in\mathcal{S}} \log (f_{\theta_{\mathcal{S}}}(y;x))\\
   -\lambda\cdot\sum_{A_i\in\mathcal{A}}\frac{1}{|A_i|}  &\sum_{x\in A_i} \sum_{y\in \mathcal{Y}} f_{\theta_{A_i}}(y;x)\log \big(f_{\theta_{A_i}}(y;x)\big)\,, 
   \end{split}
\end{equation}
where $\Theta^s=\{\theta^s_k \,|\, k \in \mathcal{K}\}$ represents the whole set of domain-specific parameters and $\lambda$ is the trade off between the cross-entropy and the entropy loss.

While \eqref{eq:loss-combined} allows to optimize the domain-specific scale and bias parameters, it does not take into account the presence of the relationship between the domains, as imposed by the graph. A way to include the graph within the optimization procedure is to modify \eqref{eq:graph-forward-std} as follows:
\vspace{-1pt}
\begin{equation}
  \label{eq:graph-forward-collaborative}
  \GBN(x,v,\mathcal{G})={\gamma}_v^\mathcal{G} \cdot\frac{x-\mu_v}{\sqrt{\sigma_v^2+\epsilon}} + {\beta}_v^\mathcal{G}\,,
\end{equation}
\vspace{-1pt}
where we have
\vspace{-1pt}
\begin{equation}
    \label{eq:training-combination-sb}
    \nu_v^\mathcal{G}=\frac{\sum_{k\in \mathcal{K}}\omega(v,k)\cdot \nu_{k}}{\sum_{k\in \mathcal{K}}\omega(v,k)}\,,
\end{equation}
%\vspace{-2pt}
for $\nu\in\{\beta,\gamma\}$.
 Basically, we use scale and bias parameters during the forward pass which are influenced by the graph edges, as described in \eqref{eq:training-combination-sb}.
 
 Taking into account the presence of $\mathcal{G}$ during the forward pass is beneficial for mainly two reasons. First, it allows to keep a consistency between how those parameters are computed at test time and how they are used at training time. Second, it allows to regularize the optimization of $\gamma_v$ and $\beta_v$, which may be beneficial in cases where a domain contains few data. While the same procedure may be applied also for ${\mu_v,\sigma_v}$, in our current design we avoid mixing them during training. This choice is related to the fact that each image belongs to a single domain and keeping the statistics separate allows to estimate them more precisely.
 
 {At test time, we initialize the domain-specific statistics and parameters of $\mathcal{T}$ given metadata $m_\set T$ using \eqref{eq:regression-metadata}, computing the forward pass of each GBN through \eqref{eq:graph-forward-collaborative}. If no metadata are available, we compute the statistics and parameters through \eqref{eq:regression-image}, performing the forward pass through \eqref{eq:graph-forward-std}. In Figure \ref{fig:method}, we sketch the behaviour of our method given $m_{\set T}$ both at training and test time.}
 
 %\massi{At test time, given once we have initialized the domain-specific statistics and parameters of $\mathcal{T}$ using either \eqref{eq:regression-metadata} or \eqref{eq:regression-image}, the forward pass of each GBN layer is computed through \eqref{eq:graph-forward-collaborative}. In Figure \ref{fig:method}, we sketch the behaviour of our method both at training and test time.}

%\vspace{-15pt}

\subsection{Model Refinement through Joint Prediction and Adaptation}
\label{sec:continuous-da}
\vspace{-4pt}
{The approach described in the previous section allows to instantiate GBN parameters and statistics for a novel domain $\mathcal{T}$ given the target metadata $m_{\set T}$. However, without any sample of $\mathcal{T}$, we have no way to assess how well the estimated statistics and parameters approximate the real ones of the target domain. This implies that we do not have the possibility to correct the parameters from a wrong initial estimates, a problem which may occur \eg if we have noisy metadata. A possible strategy to solve this issue is to exploit the images we receive at test time to refine the GBN layers. To this extent, we propose a simple strategy for performing continuous domain adaptation~\cite{hoffman2014continuous} within AdaGraph. 

Formally, let us define as $X_\mathcal{T}=\{x_1,\cdots,x_T\}$ the set of images of $\mathcal{T}$ that we receive at test time. Without loss of generality, we assume that the images of $X_\mathcal{T}$ are processed sequentially, one by one. Given the sequence $X_\mathcal{T}$, our goal is to refine the statistics $\{\mu_\mathcal{T},\sigma_\mathcal{T}\}$ and the parameters $\{\gamma_\mathcal{T},\beta_\mathcal{T}\}$ of each GBN layer as new data arrives. Following recent works~\cite{mancini2018kitting,li2018adaptive}, we continuously adapt a model to the target domain by feeding as input to the network batches of target images, updating the statistics as in standard BN. In order to achieve this, we store target samples in a buffer $M$. The buffer $M$ has a fixed size and stores the samples one by one. Exploiting the buffer, we update the target statistics as follows:
\begin{equation}
  \label{eq:online-update}
  \begin{split}
     \mu_\mathcal{T}&\longleftarrow(1-\alpha)\cdot{\mu_\mathcal{T}} + \alpha\cdot\mu_{M}\\
  \sigma^2_\mathcal{T}&\longleftarrow(1-\alpha)\cdot\sigma^2_\mathcal{T} + \alpha\cdot\frac{|M|}{|M|-1}\cdot\sigma^2_M \,,
  \end{split}
\end{equation}
where $\{\mu_M,\sigma_M\}$ are computed through \eqref{eq:graph-stats}, replacing $\mathcal{B}_v$ with $M$:
\begin{equation}
  \label{eq:online-stats}
 {\mu_M}=\frac{1}{|M|} \sum_{x\in M} x\;\;\text{and}\;\;
  {\sigma^2_M}=\frac{1}{|M|} \sum_{x\in M} (x-\mu_M)^2.
\end{equation}

While this allows to update the statistics, using \eqref{eq:online-update} does not produce any refinement on $\{\gamma_{\mathcal{T}},\beta_{\mathcal{T}}\}$. To this extent, we can easily employ the entropy term in \eqref{eq:loss-combined}:
\begin{equation}
    \label{eq:loss-online}
    L(\theta_{\mathcal{T}}) =-\frac{1}{|M|}\sum_{x \in M}\sum_{y\in \mathcal{Y}} f_{\theta_{\mathcal{T}}}(y;x)\log \big(f_{\theta_{\mathcal{T}}}(y;x)\big).
\end{equation}

To summarize, with \eqref{eq:online-update} and \eqref{eq:loss-online} we define a simple refinement procedure for AdaGraph which allows to recover from bad initialization of the predicted parameters and statistics. The update of statistics and parameters is performed together, each time the buffer is full. To avoid producing a bias during the refinement, we clear the buffer after each update step.

}

\section{Experiments}
 \subsection{Experimental setting}
 \vspace{-10pt}
\myparagraph{Datasets.}
We analyze the performance of our approach on three datasets: the \textit{Comprehensive Cars} (CompCars) \cite{yang2015large}, the \textit{Century of Portraits}  \cite{ginosar2015century} and the \textit{CarEvolution} \cite{RematasICCVWS13}.

The \textit{Comprehensive Cars} (CompCars) \cite{yang2015large} dataset is a large-scale database composed of 136,726 images spanning a time range between 2004 and 2015. As in \cite{yang2016multivariate}, we use a subset of 24,151 images with 4 types of cars (\texttt{MPV}, \texttt{SUV}, \texttt{sedan} and \texttt{hatchback}) produced between 2009 and 2014 and taken under 5 different view points (front, front-side, side, rear, rear-side). Considering each view point and each manufacturing year as a separate domain we have a total of 30 domains. As in \cite{yang2016multivariate} we use a PDA setting where 1 domain is considered as source, 1 as target and the remaining 28 as auxiliary sets, for a total of 870 experiments. In this scenario, the metadata are represented as vectors of two elements, one corresponding to the year and the other to the view point, encoding the latter as in \cite{yang2016multivariate}.

\textit{Century of Portraits} (Portraits) \cite{ginosar2015century} is a large scale collection of images taken from American high school yearbooks. The portraits are taken over 108 years (1905-2013) across 26 states. We employ this dataset in a gender classification task, in two different settings. In the first setting we test our PDA model in a leave-one-out scenario, with a similar protocol to the tests on the \textit{CompCars} dataset. In particular, to define domains we consider spatio-temporal information and we cluster images according to decades and to spatial regions (we use 6 USA regions, as defined in \cite{ginosar2015century}). Filtering out the sets where there are less than 150 images, we obtain 40 domains, corresponding to 8 decades (from 1934 on) and 5 regions (\textit{New England}, \textit{Mid Atlantic}, \textit{Mid West}, \textit{Pacific}, \textit{Southern}). We follow the same experimental protocol of the \textit{CompCars} experiments, \ie we use one domain as source, one as target and the remaining 38 as auxiliaries. We encode the domain metadata as a vector of 3 elements, denoting the decade, the latitude (0 or 1, indicating north/south) and the east-west location (from 0 to 3), respectively. %All elements of the vector are normalized between $[0,1]$. 
Additional details can be found in the supplementary material. 
In a second scenario, we use this dataset for assessing the performance of our continuous refinement strategy. In this case we employ all the portraits before 1950 as source samples and those after 1950 as target data.  

\textit{CarEvolution} \cite{yang2015large} is composed of car images collected between 1972 and 2013. It contains 1008 images of cars produced by three different manufacturers with two car models each, following the evolution of the production of those models during the years. We choose this dataset in order to assess the effectiveness of our continuous domain adaptation strategy. A similar evaluation has been employed in recent works considering online DA \cite{li2018domain}. As in \cite{li2018domain}, we consider the task of manufacturer prediction where there are three categories: \texttt{Mercedes}, \texttt{BMW} and \texttt{Volkswagen}. Images of cars before 1980 are considered as the source set and the remaining are used as target samples.

\vspace{-5pt}
\myparagraph{Networks and Training Protocols.}
To analyze the impact on performance of our main contributions we consider the ResNet-18 architecture \cite{he2016deep} and perform experiments on the Portraits dataset. In particular, we apply our model by replacing each BN layer with its AdaGraph counterpart. We start with the network pretrained on ImageNet, training it for 1 epoch on the source dataset, employing Adam as optimizer with a weight decay of $10^{-6}$ and a batch-size of 16. We choose a learning rate of $10^{-3}$ for the classifier and $10^{-4}$ for the rest of the architecture. We train the network for 1 epoch on the union of source and auxiliary domains to extract domain-specific parameters. We keep the same optimizer and hyper-parameters except for the learning rates, decayed by a factor of 10. The batch size is kept to 16, but each batch is composed by elements of a single pair year-region belonging to one of the available domains (either auxiliary or source). The order of the pairs is randomly sampled within the set of allowed ones.%(either auxiliary or source).

In order to fairly compare with previous methods we also consider Decaf features \cite{donahue2014decaf}.  In particular, in the experiments on the CompCars dataset, we use Decaf features extracted at the \texttt{fc7} layer. Similarly, for the experiments on CarEvolution, we follow \cite{li2018domain} and use Decaf features extracted at the \texttt{fc6} layer. In both cases, we apply our model by adding either a BN layer or our AdaGraph approach directly to the features, followed by a ReLU activation and a linear classifier. For these experiments %on the CompCars and CarEvolution datasets, 
we train the model on the source domain for 10 epochs using Adam as optimizer with a learning rate of $10^{-3}$, a batch-size of 16 and a weight decay of $10^{-6}$. The learning rate is decayed by a factor of 10 after 7 epochs. For CompCars, when training with the auxiliary set, we use the same optimizer, batch size and weight decay, with a learning rate $10^{-4}$ for 1 epoch. Domain-specific batches are randomly sampled, as for the experiments on Portraits.

For all the experiments we use as distance measure %for computing the edges weigths
{$d(x,y)=\frac{1}{2\cdot\sigma}\cdot||x-y||_2^2$  with $\sigma=0.1$ and set %the weight of the entropy loss 
$\lambda$ equal to $1.0$, both in the training and in the refinement stage.}
At test time, we classify each input image as it arrives, performing the refinement step after the classification. {The buffer size in the refinement phase is equal to 16 and we set $\alpha=0.1$, the same used for updating the GBN components while training with the auxiliar domains.

We implemented\footnote{The code is available at \url{https://github.com/mancinimassimiliano/adagraph}} our method with the PyTorch~\cite{paszke2017automatic} framework  and  our  evaluation  is  performed  using  a  NVIDIA GeForce 1080 Ti GTX GPU.} % and the hyper-parameters for the training with the entropy loss on the incoming target are the same adopted for the auxiliary sets.

\vspace{-4pt}
\subsection{Results}
\vspace{-4pt}
In this section we report the results of our evaluation, showing both an empirical analysis of the proposed contributions and a comparison with state of the art approaches. 

\vspace{-7pt}
\myparagraph{Analysis of AdaGraph.}
We first analyze the performance of our approach by employing the Portraits dataset. In particular, we evaluate the impact of (i) introducing a graph to predict the target domain BN statistics (\textit{\textit{AdaGraph BN}}), (ii) adding scale and bias parameters trained in isolation (\textit{AdaGraph SB}) or jointly (\textit{AdaGraph Full}) and (iii) adopting the proposed refinement strategy (\textit{AdaGraph + Refinement}). As baseline\footnote{We do not report the results of previous approaches \cite{yang2016multivariate} since the code is not publicly available.} we consider the model trained only on the source domain and, as an upper bound, a corresponding DA method which is allowed to use target data during training. In our case, the upper bound corresponds to a model similar to the method proposed in \cite{carlucci2017autodial}. %, trained with the same hyper-parameters used for the auxiliary domains}. 

The results of our ablation are reported in Table \ref{tab:portraits-PDA}, where we report the average classification accuracy corresponding to two scenarios: \textit{across decades} (considering the same region for source and target domains) and \textit{across regions} (considering the same decade for source and target dataset). The first scenario corresponds to 280 experiments, while the second to 160 tests.
As shown in the table, by simply replacing the statistics of BN layers of the source model with those predicted through our graph a large boost in accuracy is achieved ($+4\%$ in the \textit{across decades} scenario and $+2.4\%$ in the \textit{across regions} one). At the same time, estimating the scale and bias parameters without considering the graph is suboptimal. In fact there is a misalignment between the forward pass of the training phase (\ie considering only domain-specific parameters) and how these parameters will be combined at test time (\ie considering also the connection with the other nodes of the graph). %In fact, our joint training strategy for scale and bias, outperforms the separately training case in both settings (+1\% for the across decades scenario and +0.5\% for the across region one). 
Interestingly, in the \textit{across regions} setting, our full model slightly drops in performance with respect to predicting only the BN statistics. %OLDWe ascribe this behaviour to the fact that 
This is probably due to how regions 
%because the way regions 
are encoded in the metadata (\ie considering geographical location), making it difficult to capture factors (\eg cultural, historical) which can be more discriminative to characterize the population of a region or a state. However, as stated in Section \ref{sec:continuous-da}, employing a continuous refinement strategy allows the method to compensate for prediction errors. As shown in Table \ref{tab:portraits-PDA}, with a refinement step (\textit{AdaGraph + Refinement}) the accuracy constantly increases, %even in the case of "noisy" metadata, it allows to largely 
filling the gap between the performance of the initial model and our DA upper bound.

It is worth noting that applying the refinement procedure to the source model (\textit{Baseline + Refinement}) leads to better performance (about $+4\%$ in the \textit{across decades} scenario and $+2.1\%$ for \textit{across regions} one).
%{clearly demonstrating the validity of the proposed continuous parameter update}. 
More importantly, the performance of the \textit{Baseline + Refinement} method is always worse than what obtained by \textit{AdaGraph + Refinement}, % by applying the same refinement procedure starting from the model predicted by AdaGraph (\textit{AdaGraph + Refinement}), with more than 2.5\% gain in the across decades adaptation scenario. This shows 
because our model provides, on average, a better starting point for the refinement procedure.

\begin{figure*}[t]
   \includegraphics[width=1.0\textwidth]{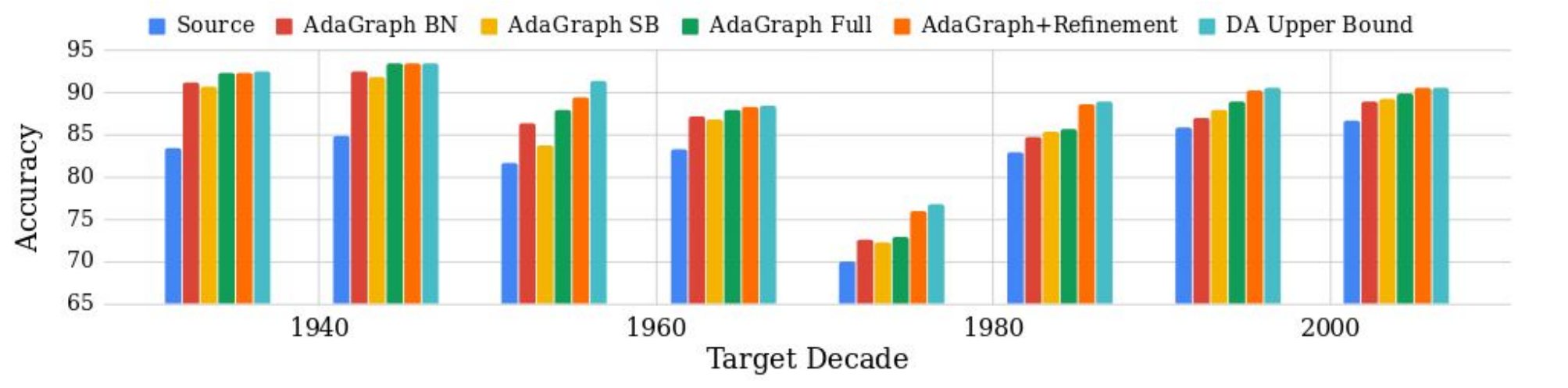}\vspace{-5pt}
   \caption{Portraits dataset: comparison of different models in the PDA scenario with respect to the average accuracy on a target decade, fixed the same region of source and target domains. The models are based on ResNet-18.}
   \vspace{-14pt}
\label{fig:ablation}
\end{figure*}

Figure \ref{fig:ablation} shows the results associated to the \textit{across decades} scenario. Each bar plot corresponds to experiments where the target domain is associated to a specific year. % corresponding to different experiments associated to different years are considered separately.  %The figure visually shows the trend described in Table \ref{tab:portraits-PDA}. 
As shown in the figure, on average, our full model outperforms both \textit{AdaGraph BN} and \textit{AdaGraph SB}, showing the benefit of the proposed graph strategy. %the prediction of BN statistics only and the isolated training of scale and bias parameters; (ii)
The results in the figure clearly also show that our refinement strategy always leads to a boost in performance. %, bringing them close to the DA upper bound.

\begin{table}[t]
			\caption{Portraits dataset. Ablation study.} 
		\centering
		\scalebox{.9}{
		\begin{tabular}{ l  | c  c} 
			\hline
			Method & Across Decades & Across Regions \\\hline
           Baseline & 82.3& 89.2\\
           AdaGraph BN & 86.3&\textbf{91.6}\\
           AdaGraph SB & 86.0&90.5\\
           AdaGraph Full & \textbf{87.0}&91.0\\\hline
           Baseline + Refinement & 86.2&91.3\\
           AdaGraph + Refinement & \textbf{88.6} &\textbf{91.9}\\
           \hline\hline
           DA upper bound &89.1&92.1 \\ \hline
        \end{tabular}
        }
        \vspace{-17pt}
        \label{tab:portraits-PDA}
\end{table}

\vspace{-8pt}
\myparagraph{Comparison with the state of the art.}
Here we compare the performances of our model with state of the art PDA approaches. We use the CompCars dataset and we benchmark against the Multivariate Regression (MRG) methods proposed in \cite{yang2016multivariate}. %, where the authors employ multivariate regression to infer the mapping between a metadata and a point in a Grassmanian manifold. 
%\eli{Ma c'è un nome migliore per imetodi di timothY??}
%Given this mapping and a metadata for the target domain, the authors propose different ways to infer the subspace in the manifold relative to the metadata itself involving \textit{Direct} kernel regression  or an \textit{Indirect} method. %The Geodesic Flow Kernel \cite{gong2012geodesic} algorithm is used to produce a model for the target domain exploiting the given source. %To compare our model with \cite{yang2016multivariate} we employ Decaf7 features \cite{donahue2014decaf} as input to the algorithm, comparably to the features used in \cite{yang2016multivariate} (\ie penultimate layer of the VGG-F model in \cite{chatfield2014return}). 

We apply our model in the same setting as \cite{yang2016multivariate} and %using one of the 30 domains as source, one as target and the other 28 remaining as auxiliaries, for a total of 870 experiments. We 
 perform 870 different experiments, computing the average accuracy (Table \ref{tab:compcars-sota}). Our model outperforms the two methods proposed in \cite{yang2016multivariate} by improving the performances of the Baseline network by $4\%$. {AdaGraph} alone outperforms the Baseline model when it is updated with our refinement strategy and target data (\textit{Baseline + Refinement}). %, demonstrating the effectiveness of our PDA approach. 
When coupled with a refinement strategy, our graph-based model further improves the performances, {filling the gap between AdaGraph and our DA upper bound}. It is interesting to note that our model is also effective when there are no metadata available in the target domain. In the table, \textit{AdaGraph (images)} corresponds to our approach when, instead of initializing the BN layer for the target exploiting metadata, we employ the current input image and a domain classifier to obtain a probability distribution over the graph nodes, as described in Section \ref{sec:graph-da}.
The results in the Table show that \textit{AdaGraph (images)} is more accurate than \textit{AdaGraph (metadata)}. 

\begin{table}[t]
			\caption{CompCars dataset \cite{yang2015large}. Comparison with state of the art.} 
		\centering
		\scalebox{.9}{
		\begin{tabular}{ l  | c } 
			\hline
			Method & Avg. Accuracy\\\hline
           Baseline \cite{yang2016multivariate}& 54.0\\
           Baseline + BN & 56.1\\
           MRG-Direct \cite{yang2016multivariate}& 58.1\\
           MRG-Indirect \cite{yang2016multivariate}& 58.2\\
           AdaGraph (metadata) & 60.1 \\
           AdaGraph (images) & \textbf{60.8} \\
           \hline
           Baseline + Refinement & 59.5\\
           AdaGraph + Refinement & \textbf{60.9}\\
        \hline\hline
        DA upper bound & {60.9}\\	\hline
        \end{tabular}
        }
        \vspace{-14pt}
        \label{tab:compcars-sota}
\end{table}
\vspace{-5pt}
\myparagraph{Exploiting AdaGraph Refinement for Continous Domain Adaptation.} In Section \ref{sec:continuous-da}, we have shown a way to boost the performances of our model by leveraging the stream of incoming target data and refine the estimates of the target BN statistics and parameters. Throughout the experimental section, we have also demonstrated how this strategy improves %and even correct the initialization of 
the target classification model, with performances close to DA methods which exploit target data during training.

In this section we show how this approach can be employed as a competitive method in the case of continuous domain adaptation \cite{hoffman2014continuous}. We consider the CarEvolution dataset and compare the performances of our proposed strategy with two state of the art algorithms: the manifold-based adaptation method in \cite{hoffman2014continuous} and the low-rank SVM strategy presented in \cite{li2018domain}. % We use the same features employed by these algorithms (Decaf6 \cite{donahue2014decaf}), in one of the scenarios of \cite{li2018domain}, using the images of cars before 1980 as sources and the rest as target. 
As in \cite{li2018domain} and \cite{hoffman2014continuous}, we apply our adaptation strategy after classifying each novel image and compute the overall accuracy. {The images of the target domain are presented to the network in a chronological order \ie from 1980 to 2013}. %To apply our model, we add a $\BN$ layer after the features and before a ReLU activation function and the final linear classifier. 
The results are shown in Table \ref{tab:ced-continuous}. While the integration of a BN layer alone leads to better performances over the baseline, our refinement strategy produces an additional boost of about 3\%. If scale and bias parameters are refined considering the entropy loss,
accuracy further increases. %there is a further increase of the performances.

We also test the proposed model on a similar task considering the Portraits dataset. % with Decaf7 features. %In this setting we take all the images before 1950 as sources and test on the rest, in chronological order. 
The results of our experiments are shown in Table \ref{tab:faces-continuous}. Similarly to what observed on the previous experiments, continuously adapting our deep model as target data become available leads to better performance with respect to the baseline. The refinement of scale and bias parameters contributes to a further boost in accuracy.

\begin{table}[t]
			\caption{CarEvolution \cite{RematasICCVWS13}: comparison with state of the art.} 
		\centering
		\scalebox{.9}{
		\begin{tabular}{ l  | c } 
			\hline
			Method & Accuracy\\\hline
           Baseline SVM \cite{li2018domain}& 39.7\\
           Baseline + BN & 43.7\\
           CMA+GFK \cite{hoffman2014continuous}& 43.0\\
           CMA+SA \cite{hoffman2014continuous}& 42.7\\
           LLRESVM \cite{li2018domain}& 43.6\\
           LLRESVM+EDA\cite{li2018domain}& 44.3\\
           Baseline + Refinement Stats& 46.5 \\
           Baseline + Refinement Full& \textbf{47.3} \\ \hline
           %\hline\hline
           %DA upper bound & \\
        \end{tabular}
        }
        \vspace{-5pt}
        \label{tab:ced-continuous}
\end{table}

\begin{table}[t]
			\caption{Portraits dataset \cite{yang2015large}: performances of the refinement strategy on the continuous adaptation scenario} 
		\centering
		\scalebox{.9}{
		\begin{tabular}{ l  | c c c } 
			\hline
			Method & Baseline & Refinement Stats & Refinement Full\\\hline
           Accuracy& 81.9& 87.3 & \textbf{88.1}\\ \hline
        \end{tabular}
        }
        \vspace{-15pt}
        \label{tab:faces-continuous}
\end{table}

\section{Conclusions}
We present the first deep architecture for Predictive Domain Adaptation. 
%In this setting we can only use metadata to predict a model for the target domain. In order 
We leverage metadata information to build a graph where each node represents a domain, while the strength of an edge models the similarity among two domains according to their metadata. We then propose to exploit the graph for the purpose of DA and we design novel domain-alignment layers. 
%By initializing each domain available at training time with specific BN layers, at test time we propagate the parameters of those layers to the target model by using only the edges connecting the target domain to the others.
This framework yields the new state of the art on standard PDA benchmarks.
We further present an approach to exploit the stream of incoming target data such as to refine the target model. %Our strategy is based both on the continuous update of the BN statistics and on the refinement of the scale and bias parameters through the entropy loss. %This allows to both increase the performances and achieve higher robustness to wrong parameters prediction and noisy metadata. 
We show that this strategy itself is also an effective method for continuous DA, outperforming state of the art approaches. %in this setting. % of the parameters for the target domain further increase the robustness of the model to wrong parameter prediction by
 Future works will explore methodologies to incrementally update the graph %as new domains are encountered 
 and to automatically infer relations among domains, even in the absence of metadata.

{\small
\bibliographystyle{ieee}
\bibliography{main}
}
\appendix

\newpage
\section{Supplementary Material}
\subsection{Metadata Details}
\myparagraph{CompCars.} For the experiments with the CompCars dataset \cite{yang2015large}, we have two domain information: the car production year and the viewpoint. We encode the metadata through a 2-dimensional integer vector where the first integer encodes the year of production (between 2009 and 2014) and the second the viewpoint. While encoding the production year is straightforward, for the viewpoint we use the same criterion adopted in \cite{yang2016multivariate}, \ie we encode the viewpoint through integers between 1-5 in the order: \textit{Front}, \textit{Front-Side}, \textit{Side}, \textit{Rear-Side}, \textit{Rear}.

\myparagraph{Portraits.} For the experiments with the Portraits dataset \cite{ginosar2015century}, we have again two domain information: the year and the region where the picture has been taken. To allow for a bit more precise geographical information we encode the metadata through a 3-dimensional integer vector.

As for the CompCars dataset, the first integer encodes the decade of the image (8 decades between 1934 and 2014), while the second and third the geographical position. For the geographical position we simplify the representation through a coarse encoding involving 2 directions: est-west (from 0 to 1) and north-south (from 0 to 3). In particular we assign the following value pairs ([north-south, east-west]): \textit{Mid-Atlantic} $\rightarrow[0,1]$, \textit{Midwestern} $\rightarrow[0,2]$, \textit{New England} $\rightarrow[0,0]$, \textit{Pacific} $\rightarrow[0,3]$ and \textit{Southern} $\rightarrow[1,1]$. Each component of the vector has been normalized in the range 0-1.

\subsection{ResNet-18 on CompCars}
Here we apply \textit{AdaGraph} to the ResNet-18 architecture in the CompCars dataset \cite{yang2015large}. As for the other experiments, we apply \textit{AdaGraph} by replacing each BN layer of the network with its GBN counterpart. 

The network is initialized with the weights of the model pretrained on ImageNet. We train the network for 6 epochs on the source dataset, employing Adam as optimizer with a weight decay of $10^{-6}$ and a batch-size of 16. The learning rate is set to $10^{-3}$ for the classifier and $10^{-4}$ for the rest of the network and it is decayed by a factor of 10 after 4 epochs. We extract domain-specific parameters by training the network for 1 epoch on the union of source and auxiliary domains, keeping the same optimizer and hyper-parameters. The batch size is kept to 16, building each batch with elements of a single pair
production year-viewpoint belonging to one of the domains available during training (either auxiliary or source). %For \textit{DA upper bound}, we again consider the target specific parameters extracted in the same way we extract the parameters of each auxiliary domain.

{The results are shown in Table \ref{tab:compcars-resnet}. As the table shows, \textit{AdaGraph} largely increases the performance of the \textit{Baseline} model}. Coherently with previous experiments, our refinement strategy is able to further increase the performances of \textit{AdaGraph}, filling almost entirely the gap with the DA upper bound.

\begin{table}[t]
			\caption{CompCars dataset \cite{yang2015large}. Results with ResNet-18 architecture.} 
		\centering
		\scalebox{1.}{
		\begin{tabular}{ l  | c } 
			\hline
			Method & Avg. Accuracy\\\hline
           Baseline& 56.8\\
           AdaGraph & \textbf{65.1} \\
           %AdaGraph (images) &  \\
           \hline
           Baseline + Refinement & 65.3 \\
           AdaGraph + Refinement & \textbf{66.7} \\
        \hline\hline
        DA upper bound & 66.9\\	\hline
        \end{tabular}
        }
        \label{tab:compcars-resnet}
\end{table}

\subsection{Performances vs Number of Auxiliary Domains}
\begin{figure}[ht!]
  \begin{subfigure}[t]{1.\columnwidth}
    \includegraphics[width=\columnwidth,height=0.5\columnwidth]{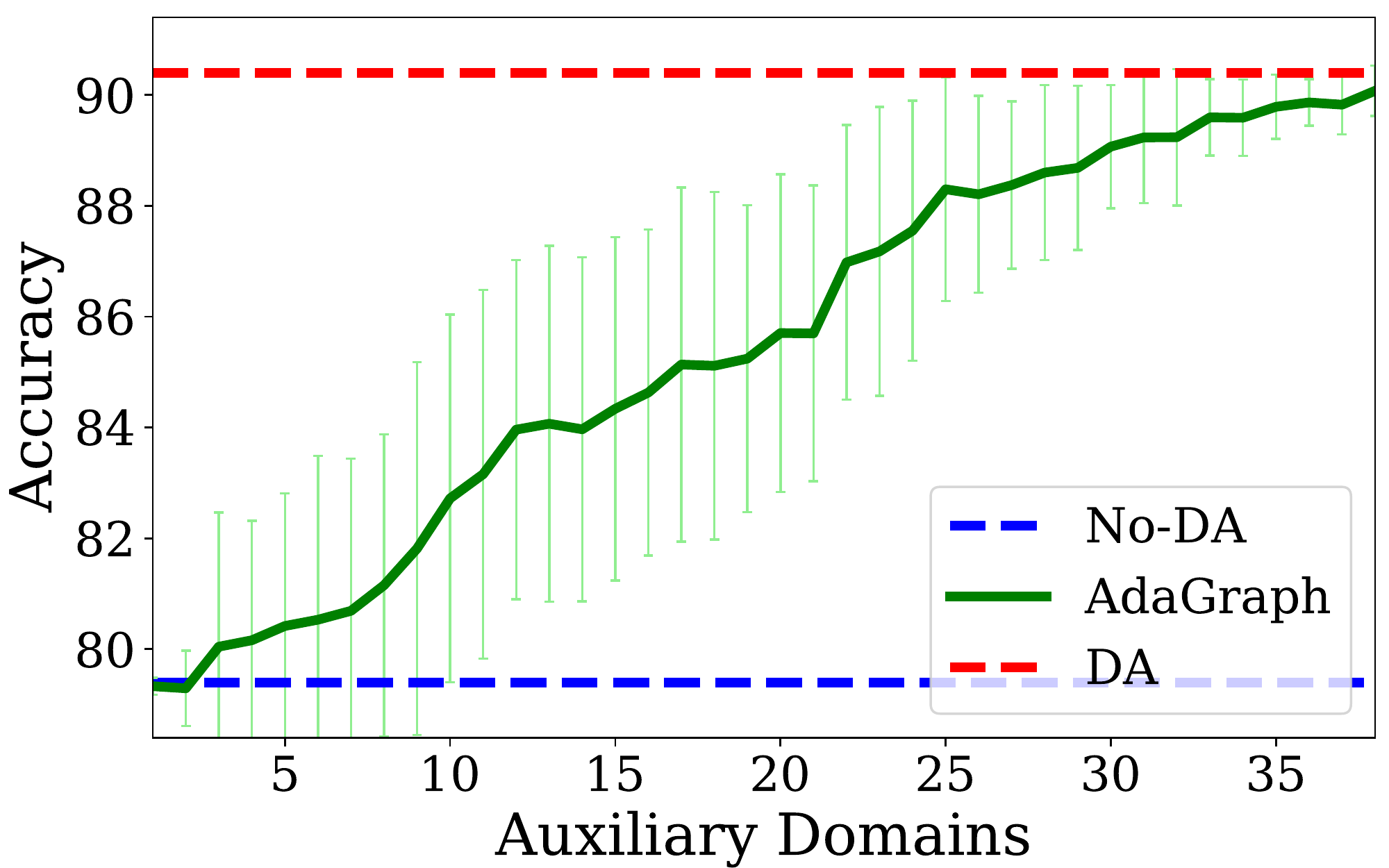}
    \caption{From 1954 Mid-Atlantic to 1984 Pacific.}
    \label{fig-a-faces}
  \end{subfigure}\\
  \begin{subfigure}[t]{1\columnwidth}
    \includegraphics[width=1.\columnwidth,height=0.5\columnwidth]{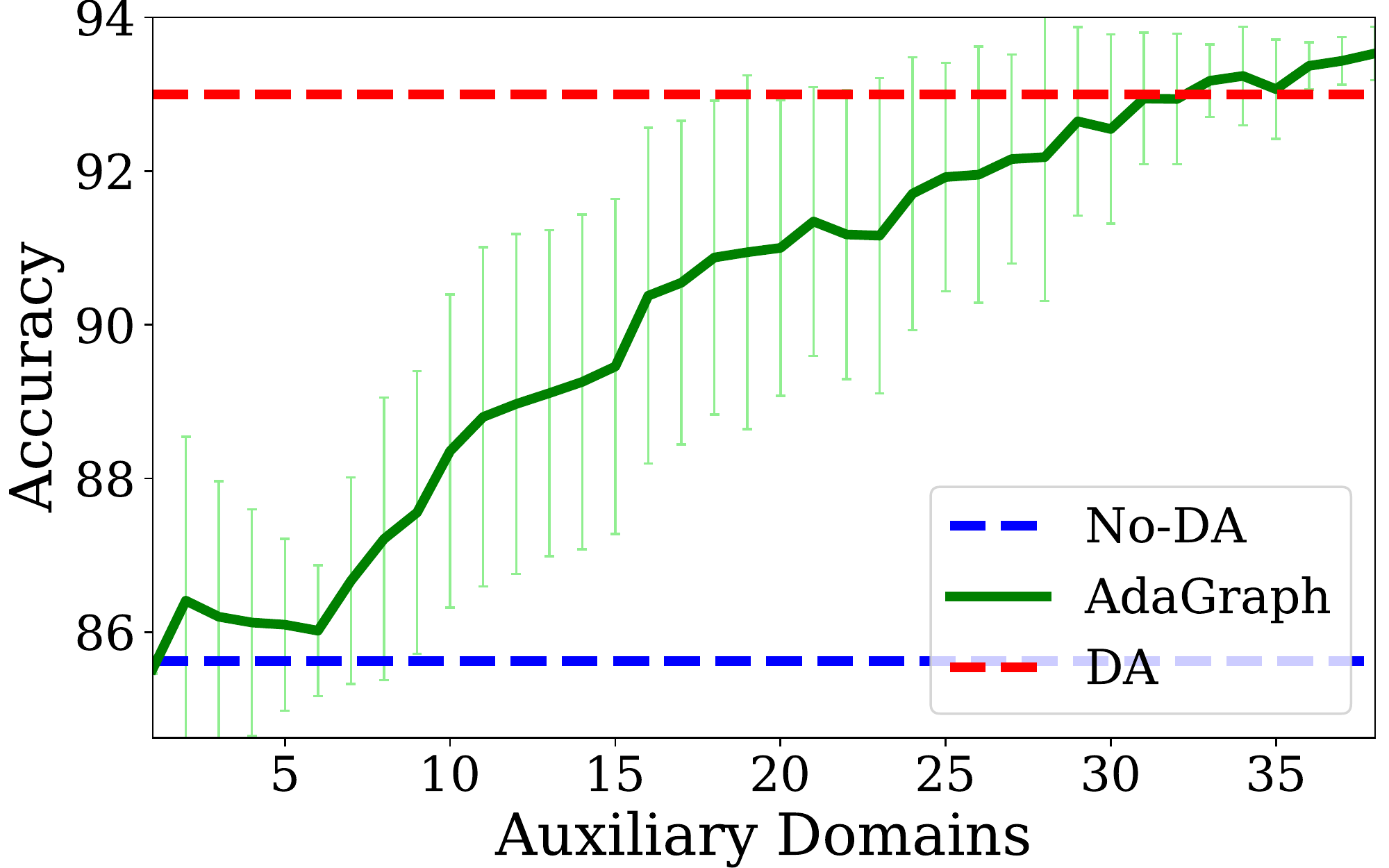}
    \caption{From 2004 Midwestern to 1944 Southern.}
    \label{fig-b-faces}
  \end{subfigure}\\
    \begin{subfigure}[t]{1\columnwidth}
    \includegraphics[width=\columnwidth,height=0.5\columnwidth]{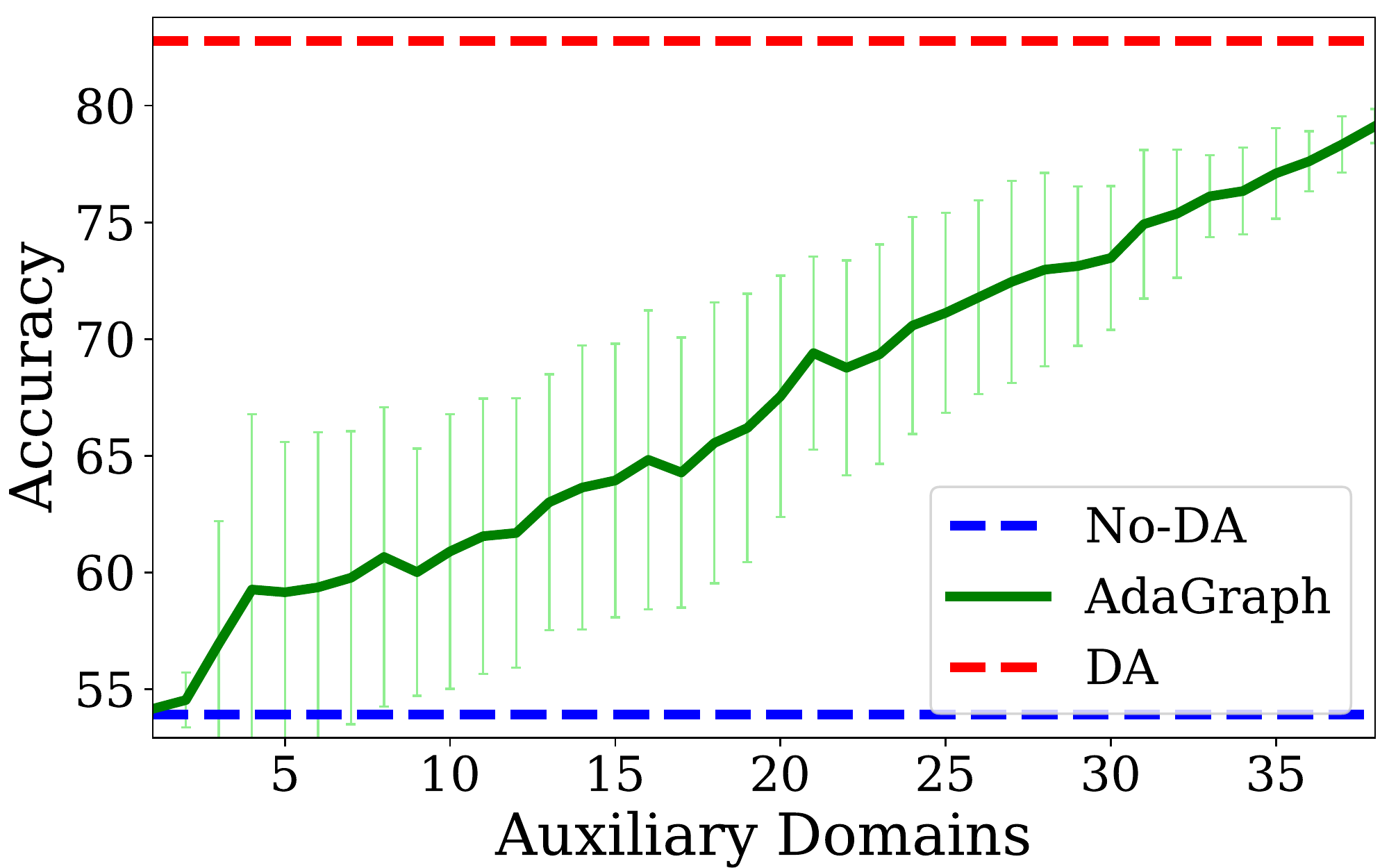}
    \caption{From 1974 Mid-Atlantic to 1994 New England.}
    \label{fig-c-faces}
  \end{subfigure}
  \caption{Portraits dataset: performances of \textit{AdaGraph} with respect to the number of auxiliary domains available for different source-target pairs. The years reported in the captions indicate the starting year of source and target decades.} 
  \label{fig:accuracy-vs-domains-portraits}
\end{figure}

%\begin{figure}[t]
%\begin{center}
 %  \includegraphics[width=1.0\columnwidth]{imgs/plot-1954MA-1984Pacific.pdf}
%\end{center}
 %  \caption{1954 MA to 1984 Pacific}
%\label{fig:accuracy-vs-domains}
%\end{figure}

%\begin{figure*}[t]
  %\begin{subfigure}[t]{0.32\textwidth}
    %\includegraphics[width=\textwidth]{imgs/plot-2009-1_2013-5.pdf}
    %\caption{From 2009 Front to 2013 Rear.}
   % \label{fig-a-cars}
  %\end{subfigure}\hfill
  %\begin{subfigure}[t]{0.32\textwidth}
    %\includegraphics[width=\textwidth]{imgs/plot-2012-3_2009-2.pdf}
    %\caption{From 2012 Side to 2009 Front-side.}
   % \label{fig-b-cars}
  %\end{subfigure}\hfill
    %\begin{subfigure}[t]{0.32\textwidth}
    %\includegraphics[width=\textwidth]{imgs/plot-2010-2_2011-4.pdf}
    %\caption{From 1974 Mid-Atlantic to 1984 New %England.}
  %  \label{fig-c-cars}
  %\end{subfigure}
  %\caption{Portraits dataset: performances of %\textit{AdaGraph} with respect to the number of %auxiliary domains available for different %source-target pairs.} 
 % \label{fig:accuracy-vs-domains-compcars}
%\end{figure*}

In this section, we analyze the impact of varying the number of available auxiliary domains on the performances of our model. We employ the ResNet-18 architecture on the Portraits dataset, with the same setting and set of hyperparameters described in the experimental section. However, differently from the previous experiments, we vary the number of available auxiliary domains, from 1 to 38. % at their impact on the performances of the model. 
We repeat the experiments 20 times, randomly sampling the available auxiliary domains each time.

The results are shown in Figure \ref{fig:accuracy-vs-domains-portraits}. As expected, increasing the number of auxiliary domains leads to an increase in the performance of the model. In general, as we have more than 20 domains available, the performance of our model are close to the DA upper bound. While these results obviously depend on the relatedness between the auxiliary domains and the target, the plots show that having a large set of auxiliary domains may not be strictly necessary for achieving good performances.

\end{document}